# COVID-19 Twitter Sentiment Classification Using Hybrid Deep Learning Model Based on Grid Search Methodology

[1,*]Jitendra V. Tembhurne, [2]Anant Agrawal, [3]Kirtan Lakhotia

[1]Department of Computer Science & Engineering, Indian Institute of Information Technology, Nagpur, India
[2]Department of Computer Science & Data Science, Shri Ramdeobaba College of Engineering & Management, Nagpur, India
[3]Department of Information Technology, Institute of Engineering and Technology, Devi Ahilya Vishwavidyalaya, Indore, India

[1]jtembhurne@iiitn.ac.in, [2]anantagrawal0510@gmail.com, [3]kirtanlakhotia@gmail.com

**Abstract**

In the contemporary era, social media platforms amass an extensive volume of social data contributed by their users. In order to promptly grasp the opinions and emotional inclinations of individuals' regarding a product or event, it becomes imperative to perform sentiment analysis on the user-generated content. Microblog comments often encompass both lengthy and concise text entries, presenting a complex scenario. This complexity is particularly pronounced in extensive textual content due to its rich content and intricate word interrelations compared to shorter text entries. Sentiment analysis of public opinion shared on social networking websites such as Facebook or Twitter has evolved and found diverse applications. However, several challenges remain to be tackled in this field. The hybrid methodologies have emerged as promising models for mitigating sentiment analysis errors, particularly when dealing with progressively intricate training data. In this article, to investigate the hesitancy of COVID-19 vaccination, we propose eight different hybrid deep learning models for sentiment classification with an aim of improving overall accuracy of the model. The sentiment prediction is achieved using embedding, deep learning model and grid search algorithm on Twitter COVID-19 dataset. According to the study, public sentiment towards COVID-19 immunisation appears to be improving with time, as evidenced by the gradual decline in vaccine reluctance. Through extensive evaluation, proposed model reported an increased accuracy of 98.86%, outperforming other models. Specifically, the combination of BERT + CNN + GS yield the highest accuracy, while the combination of GloVe + Bi-LSTM + CNN + GS follows closely behind with an accuracy of 98.17%. In addition, increase in accuracy in the range of 2.11% to 14.46% is reported by the proposed model in comparisons with existing works. Thus, the combination of embedding and grid search increases the accuracy by using deep learning models.

## 1. Introduction

In today's interconnected world, individuals and organizations generate an unprecedented volume of textual content daily. This content spans across news articles, social media posts, customer feedback, and product reviews, etc. Understanding the emotions and opinions conveyed within this textual data is invaluable for various domains, including marketing, customer service, finance, politics, and public opinion research. Sentiment analysis provides a structured approach to deciphering this emotional content, enabling better decision-making, improved customer experiences, and enhanced strategic planning [39].

Sentiment analysis (SA), at its core, is the process of automatically determining the emotional tone or opinion expressed in the text or multimedia contents. It targets to recognize the sentiment categories i.e. positive, negative, or neutral, and may go beyond these categories to capture nuances such as joy, sadness or anger. The primary objectives of SA includes understanding public sentiment pertaining to particular topic, product, or entity, tracking the trends of sentiment over the time, and extracting actionable insights to inform decision-making processes. The applications of sentiment analysis are multifaceted and continually expanding [40-42]. In business and marketing [44], companies use sentiment analysis to gauge customer opinions, track brand reputation, and fine-tune advertising campaigns. In politics [45], sentiment analysis helps in monitoring public sentiment towards candidates, policies, and elections. In healthcare [43], it aids in analyzing patient feedback and identifying potential issues in healthcare services. Furthermore, sentiment analysis is instrumental in financial markets, social media monitoring, and customer service optimization. Sentiment analysis encompasses a variety of methodologies and techniques, spanning from rule-based approaches to machine learning (ML) and deep learning methods. The Rule-based methods depends on predefined sets of rules and linguistic patterns to classify sentiment. In contrary, ML and DL models employ algorithms to automatically learn sentiment patterns from labelled training data. Techniques namely Natural Language Processing (NLP), the feature engineering play a vital role in extracting meaningful features from text, making it suitable for SA.

Traditional sentiment analysis has relied on several well-established algorithms and techniques. These methods, often based on rule-based and statistical approaches, have been instrumental in analyzing sentiment in text data. Some of the traditional algorithms utilized for sentiment analysis such as Lexicon-Based Methods (SentiWordNet and AFINN lexicon), k-nearest neighbour (KNN), Naive Bayes, Maximum Entropy, Support Vector Machines (SVM), N-grams, and Bag-of-Words (BoW), etc. SA of students' comments on Twitter were processed by Aung et al. [1] wherein lexicon based method was adopted i.e. entity-level SA. Initially, lexicon based method was applied for entity-level SA, achieved increase in precision, but lesser recall. Further, additional tweets are identified automatically which consist some opinion as the result of aforementioned method, thus, some improvement in recall is noticed. Progressively, Dey et al. [2] experimented two supervised ML algorithms namely Naïve Bayes and KNN, and compared overall accuracy, precision and recall values of both the algorithms. It was observed that Naïve Bayes outperforms the KNN for movie reviews but both the algorithms reported lesser accuracy for hotel reviews, where the accuracy were almost same.

In latest years, the realm of SA has witnessed a remarkable transformation through the widespread adoption of deep learning models. This adoption has not only demonstrated the immense potential but also propelled sentiment analysis to new heights. The core foundation of DL lies in the utilization of deep neural networks (DNNs). Deep learning models excel at the task of progressively transforming raw data into increasingly abstract and hierarchical representations by employing a series of nonlinear transformations. This capability enables deep models to acquire intricate functional features through the amalgamation of multiple layers of transformation. Consequently, in contrast to traditional ML models, DL proves to be more adept in handling task like sentiment classification, thereby enhancing model's overall performance. Within the realm of DNN methods, notable approaches encompass - Convolutional Neural Network (CNN), Recurrent Neural Network (RNN) and its variants.

While individual ML methods exhibit varying levels of reliability within specific domains, it's important to recognize that each deep learning approach possesses its unique strengths and weaknesses. The CNNs are effective at capturing local sentiment cues in text. However, it struggle with accounting for long-range dependencies and text order. On the other hand, Long Short-Term Memory (LSTM), a temporal recursive neural network, is skilled at processing sequential data and is suitable for tasks with extended time intervals. While LSTM integrates text information sequentially, it only considers forward information, missing out on context from the other direction. To address this limitation, Bidirectional LSTM (BiLSTM) was introduced to incorporate both forward and backward context, enhancing predictive performance. Nevertheless, BiLSTM can be computationally expensive due to high input dimensions but pooling the choice between LSTM and CNN hinges on the specific needs of a given task. The LSTM shines when dealing with lengthy textual inputs, even though it demands greater computational resources, while CNN stands out for its efficiency in tasks that require fewer hyperparameter adjustments and less manual oversight.

Some researchers have suggested the idea of using more than one method (like combining two or more) to acquire the best of both and overcome the weaknesses of each method by working together. Alfrjani et al. [3] developed the hybrid Semantic Knowledgebase (SK) ML technique for opinion mining at domain feature level and, further, overall opinions were classified based on multi-point scale. The hybrid SK technique evaluates the reviews' at feature level, and structured information associated with expressed opinions belongs to defined domain features are generated. The improved recall and precision is reported by hybrid SK technique, and hence better classification accuracy is achieved on augmented dataset with semantic features.

Rehman et al. [4] proposed a DL model consisting very deep CNN layers and LSTM namely, hybrid CNN-LSTM model for SA task. Authors used Word-to-Vector (Word2Vec) embedding, initially. Here, CNN model efficiently extracts the spatial features, and LSTM model captured long-term dependencies amongst word sequences. Afterword, embedding is performed on combined features that are extracted by CNN (conv-layer + global max-pooling-layer) consisting long term dependencies. The hybrid CNN-LSTM model resulted in better performance metrics compared with traditional ML and DL techniques.

Due to the ease of social media, all are freely expressing their opinions. Anti-vaccination activities impeded attempts to stop or slow the spread of the COVID-19 pandemic by using social media to persuade people to accept vaccines at lower rates [50]. The public's confidence in the continuation of vaccination campaigns can be weakened by untrue assertions about the Covid-19 vaccine that have been making the rounds online since the start of the pandemic [51]. The world's health is at risk because of this. Although online rumours and conspiracy theories may fuel vaccination scepticism, we contend that tracking Covid-19 related data associated with social media platform can aid in tracking vaccine statistics and lessening its impact [52]. Additionally, a successful vaccine campaign approach can be aided by a knowledge of public viewpoints from behaviour change principles [53].

Subsequently, the key objective of this study is to build an optimized framework that integrates different deep learning architectures to enhance sentiment analysis accuracy. By combining different deep learning models, the aim is to deal with the limitations of individual models and utilize the strengths of each to improve overall sentiment analysis capabilities, ultimately surpassing state-of-the-arts performance benchmark in sentiment analysis by leveraging the advantages of hybrid deep learning models. The motivation behind employing hybrid deep learning models lies in their potential to outperform single models, resulting in more precise sentiments. The future scope presented by Qorib et al. [5] is the main motivation of this work where limited accuracy of individual model is reported, further, authors designed the hybrid ML models which obtained the accuracy of 96.75%. Thus, we decided to develop the hybrid DL model using efficient word embedding and CNN, BiLSTM and Grid Search algorithm. Despite the resource-intensive nature of these models, this study evaluates eight different models with varied hyperparameter combinations, providing valuable insights for determining the most accurate prediction i.e. correct sentiment.

Sentiment analysis is a critical field with wide-ranging implications across industries. Its ability to distil emotions and opinions from textual data empowers decision-makers to better understand public sentiment and harness it for strategic advantage. This work explores the nuances of SA, from its methodologies to its diverse applications and ongoing challenges, shedding light on the exciting possibilities it holds for the future.

The remaining sections of the article are organized as follows: Section 2, reviewed of related research. Section 3 outlines the chosen method, Section 4 presents the results analysis on the experimentation, and finally, we concludes in Section 5.

## 2. Literature Study

The objective of this work is to develop a highly optimized and efficient sentiment analysis model with a focus on improving accuracy. We have conducted a thorough examination of the methodologies proposed and implemented in prior studies, which are discussed in this section.

Qorib et al. [5] tested three simulations -1) word stemming, 2) word lemmatization, and 3) combination of stemming and lemmatization. Lemmatization produced good results than the stemming in text preprocessing. The model's performance experienced a slight enhancement when stemming and lemmatization are combined. Sentiment computation using TextBlob reported the highest performance for some models. Authors, claimed that TextBlob is a better classifier compared to VADER (Valence Aware Dictionary for Sentiment Reasoning) and Azure sentiment computations. Highest model performance with precision of 96.92% was achieved for TextBlog with Term Frequency-Inverse Document Frequency (TF-IDF) with LinearSVC classifier. Subsequently, during COVID-19 pandemic breakout, the SA were performed on COVID-19 dataset [18-23]. In Kaur et al. [6], authors introduced a hybrid Heterogeneous SVM (H-SVM) for SA on Twitter data concerning the COVID-19 pandemic. The classification is performed by using RNN and SVM, and better classification accuracy was achieved.

Khan et al. [7] investigated ML methods on US-based COVID-19 tweets to classify sentiments into positive, negative, or neutral. Here, TF-IDF and BoW features were utilized and performance of different ML models were observed for sentiment classification. The Gradient Boosting Machine (GBM) outperforms, other ML models, and features from BoW and TF-IDF offers better F1-score. If model's performance comparison using TF-IDF and BoW is performed then TF-IDF's results are better than BoW. The highest accuracy of 96% and recall of 100% is achieved by GBM.

Recent analysis show that deep learning models tends to outperform traditional sentiment analysis model [24-29]. Dang et al. [8] experimented on deep NN, CNN and LSTM models alternatively with word embedding and TF-IDF on eight different dataset and highest accuracy of 90.45% was obtained using word embedding and RNN on Tweets Airline dataset. Further, to analyse sentiment on Twitter text comments, which is a mix of long text and short text, Xiaoyan et al. [9] presented GloVe-CNN-BiLSTM model. The GloVe embedding was utilized initially, to get feature vector, then CNN-BiLSTM is applied to feature learning from online tweets i.e. short-text, long-text, and complete-text, respectively. The accuracy of 95.09%, 95.60%, and 95.65% was achieved on long-text, short-text, and complete-text, respectively by GloVe-CNN-BiLSTM model. Priyadarshini and Cotton [10] presented the performance analysis of DL models with suitable hyperparameter tuning, the comparative analysis shows that the LSTM–CNN–GS outperforms better as compared to baseline algorithms.

Transfer learning has recently found success in the field of SA, it involves training the lower network layers on abundant supervised datasets such as Bidirectional Encoder Representations from Transformers (BERT) [11] and XLNET [12] - notable instances of the BERT and its variation [11-13], here, BERT and its variants were employed for SA. Talaat [13] proposed hybrid RoBERTa (i.e. BERTBase) and DistilBERT (i.e. BERTMini) with BiLSTM and Bidirectional Gated Recurrent Unit (BiGRU) to achieve better results. Eight different hybrid models were developed, and DistilBERT-GLG achieved 1.84% more accuracy over DistilBERT without emoji in Apple dataset and only 0.24% more accuracy over DistilBERT on Airline Dataset. For RoBERTa, the models with BiGRU layers achieved better performance than others, especially for large and small datasets. The accuracy enhancement is recorded for BiGRU+DistilBERT and BiGRU+RoBERTa.

Dang et al. [14] conducted rigorous experiments on SVM, CNN and LSTM, and proposed four models i.e. CNN+LSTM+ReLU, CNN+LSTM+SVM, LSTM+CNN+ReLU and LSTM+CNN+SVM, respectively. The model performance was observed on Word2vec and pretrained BERT model, and accuracy of 92.9% and 93.4% is reported on IMDB and Tweets Airline datasets, respectively.

The key findings from the aforementioned literature indicates that the limitations in sentiment prediction due to single method and limited classification accuracy were reported by single method. In addition, previous research works lacks in 1) computation methods for classification, 2) limited datasets for vaccine hesitancy, and 3) limited classification results for COVID-19 vaccination. Moreover, it is found that hybrid DL models outperform both single deep learning models and traditional ML methods. Further, utilizing word embeddings within these models, as opposed to TF-IDF, consistently yields superior results. Subsequently, among the various word embeddings techniques, BERT and GloVe demonstrated enhanced performance in comparison with FastText and Word2Vec. Additionally, Grid Search for hyperparameter optimization was found to be an effective approach for achieving more accurate predictions. The further proposed works [30-36] offers additional evidence and insights on hybrid models. We also presented the summary based on hybrid models with respective to accuracy in Table 1.

**Table 1:** Sentiment analysis based on hybrid approach.

| Ref. | Methodology | Accuracy |
| --- | --- | --- |
| Qorib et al. [5] | TextBlog+TF-IDF+LinearSVC | 96.75% |
| Kaur et al. [6] | MKH+SVM | 96.30% |
| Khan el al. [7] | GBM ensemble | 96.00% |
| Dang et al. [8] | Word embedding+RNN | 90.45% |
| Xiaoyan et al. [9] | GloVe+CNN+BiLSTM | 95.65% |
| Priyadarshini and Cotton [10] | LSTM+CNN+GS | 96.00% |
| Dang et al. [14] | CNN+LSTM | 93.44% |

## 3. Material and Methods

In this work, we evaluate the efficacy of eight hybrid DL models to enhance the performance of SA technique. Proposed methodology primarily revolves around three key components – 1) selecting the dataset, 2) constructing feature vectors, and 3) developing hybrid methods to create a suitable SA solution. These models are developed to forecast the polarity of sentiment in textual content and categorize it based on its polarity. The proposed system is presented in Fig. 1 comprising – feature representation, feature learning, and sentiment classification with extensive hyperparameters tuning by grid search algorithm.

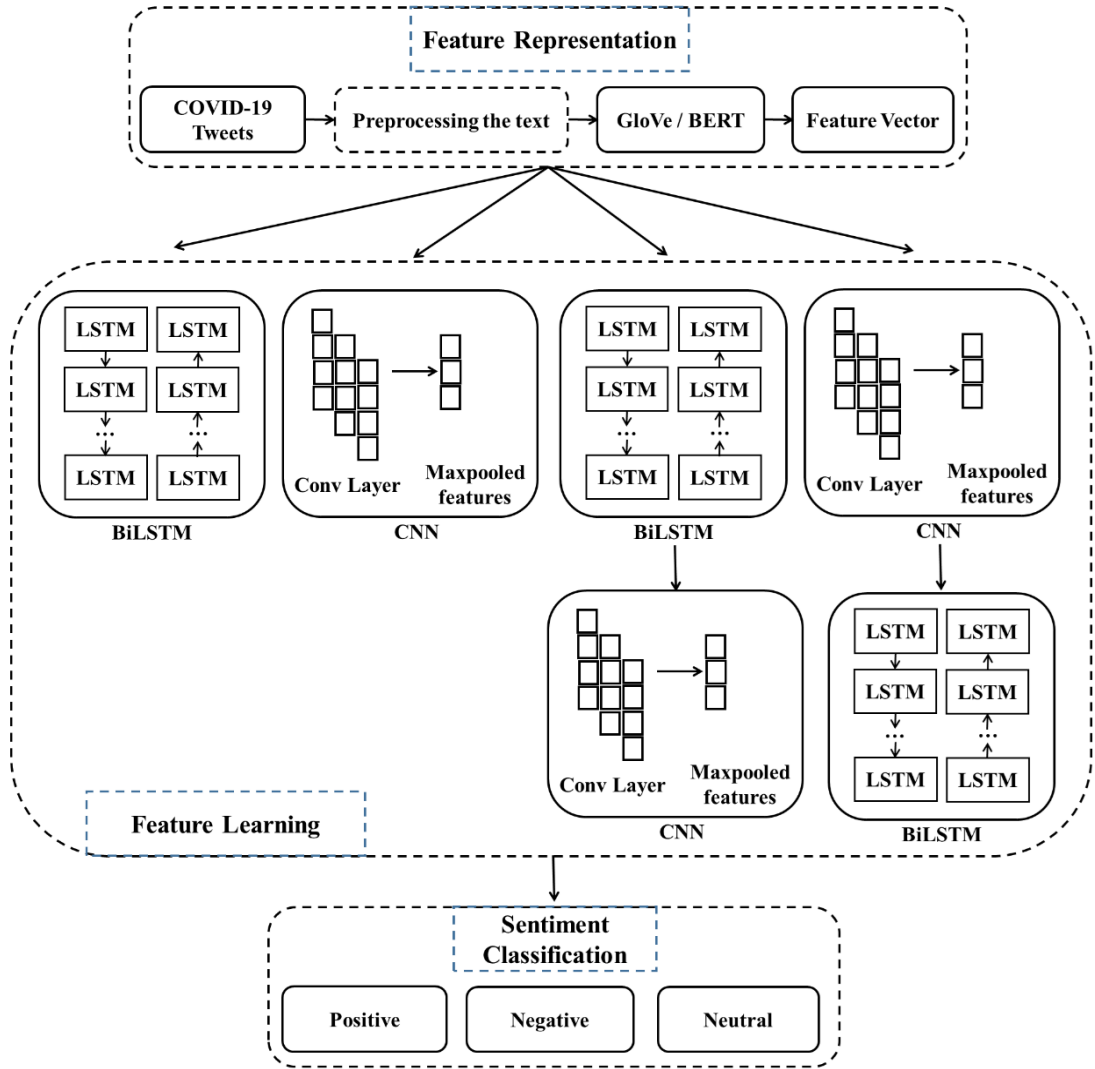

**Figure 1:** Proposed System.

### 3.1 Datasets

The evaluation of hybrid models involved the utilization of a Twitter COVID-19 dataset [15]. This dataset encompasses tweets spanning from different time frames: April to June 2020 (Dataset - A), August to October 2020 (Dataset - B), and April to June 2021 (Dataset - C). Dataset - A comprises 143,902 tweets, Dataset - B comprises 120,508 tweets, and Dataset - C comprises 147,474 tweets. The distribution of positive, negative, and neutral comments belongs to different datasets is highlighted in Table 2, moreover, Table 3 shows the sample text from the dataset.

**Table 2:** Different Datasets.

| Tweets | Dataset - A | Dataset - B | Dataset - C |
|---|---|---|---|
| Positive | 46,124 | 36,479 | 44,756 |
| Negative | 40,191 | 33,035 | 36,400 |
| Neutral | 57,587 | 50,994 | 66,318 |
| Total | 143,902 | 120,508 | 147,474 |

**Table 3:** Sample Tweets.

| Tweets | Labels |
|---|---|
| Kisumu Governor Nyong'o orders all restaurants shut in the county for failing to adhere to MoH measures against Covid-19 | *neg* |
| India backs 62 nation coalition calling for WHO to investigate China unleashing the coronavirus on the world | *neu* |
| We've been actively looking for ways to support our customers as the need for virtual communication increases | *pos* |

### 3.2 Preprocessing and Feature Vector Building

Preprocessing of the data is one of the most crucial step while feeding the data to the ML and deep learning models. Raw data must be cleaned before feeding to the model for efficient training and produce more effective results. We have used Pandas, Numpy and NLTK libraries for data cleaning and utilization purposes. The steps taken while data cleaning are mentioned in Fig. 2.

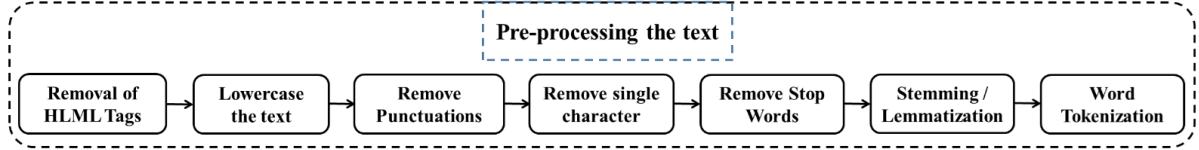

**Figure 2:** Preprocessing of text

After cleaning the text of reviews we have to also map sentiment from '*pos*', '*neu*' and '*neg*' to 0, 1 and 2, respectively. Then we have to apply one hot encoding to feed the sentiment to the model. Classification of sentiment can be conducted at 3 different levels of analysis: document-level, sentence-level, and aspect-level or feature-level. Here, we employed document-level SA using word-embeddings on Twitter COVID-19 datasets. The current landscape of word vectorization techniques; the two widely used approaches are global matrix decomposition, exemplified by Latent Semantic Analysis (LSA), and another relies on local context windows, such as skipgram model utilized in Word2Vec. Each approach has its distinct advantages and limitations. LSA leverages statistical facts for SA but tends to perform poorly in tasks related to lexical analogy. On the other hand, Word2Vec excels in lexical analogy tasks but is constrained by its focus on local context windows, making it difficult to efficient use of global lexical co-occurrence statistics.

Subsequently, in proposed model, we employed word-embeddings, specifically BERT and GloVe, to construct feature vector. Here, we utilize a fixed length, denoted as '*d*' for each dataset. This fixed length serves as a standard for processing text samples. For samples that are shorter than this fixed length '*d*' we append zeros at the end of vector representation. Conversely, for longer samples than '*d*', we truncate the excess data from the end. However, it's crucial to select an appropriate '*d*' to minimize data truncation while maintaining classification process efficient.

In this study, we work with tweet and review datasets. For tweets, we set '*d*' closed to maximum length of tweets, which is typically 100 characters. This choice is reasonable because tweets have a limited length due to platform constraints. While it's possible to use a single fixed length '*d*' for both tweets and reviews, doing so would be suboptimal. If we chose '*d*' to be longer, it would result in inefficient memory usage. Conversely, if '*d*' are smaller, valuable review data would be lost. Hence, adapting '*d*' based on dataset characteristics is a more practical approach. This method of selecting '*d*' for different datasets is consistent ensuring that we strike an appropriate balance between memory efficiency and data preservation.

### 3.3 GloVe

The GloVe [48], a hybrid approach, combines the strengths of both LSA and Word2Vec by integrating global statistical information with local context windows. This results in more effective word vectorization, achieving a balance between global and local context.

GloVe embedding is represented using Eq. (1):

$$J = \sum_{i,j}^{N} f(X_{i,j})(V_i^T V_j + b_i + b_j - \ln(X_{i,j}))^2 \qquad (1)$$

The GloVe model utilizes a formula based on a cooccurrence matrix, denoted as $X$, capturing the occurrences of words $i$ and $j$ within a specified context window. The elements of this matrix, $X_{i,j}$, represent the frequency of occurrence for words $i$ and $j$ together within the window, typically of size 5 to 10. $V_i$ and $V_j$ denote the word vectors for words $i$ and $j$ respectively, while $b_i$ and $b_j$ represent deviation terms. $N$ represents dimension of cooccurrence matrix $N \times N$. The weight function, denoted as $f$, plays a key role in the model. It assure that cooccurrences with a count of 0 have a weight of 0. Moreover, it maintains continuity and non-decrementality, meaning that as the cooccurrence count increases, the weight assigned by $f$ does not decrease. However, to prevent overemphasis on highly frequent cooccurrences, $f(x)$ assigns relatively smaller values in such cases. The $f(x)$, weight function is represented as:

$$f(x) = \begin{cases} (x/x_{\max})^\alpha, & \begin{cases} x < x_{max} \\ x \geq x_{max} \end{cases} \end{cases} \qquad (2)$$

### 3.4 BERT

Google's released BERT [11] model in 2018, marked a significant milestone in NLP, BERT has demonstrated remarkable performance in eleven classic NLP tasks, making it a vastly utilized after word-vector technique. However, it's crucial to recognize that practical applications of BERT still face some unresolved challenges, which researchers are actively exploring through further experimentation and study. The BERT model architecture is represented in Fig. 3, wherein word embedding is applied on the input, followed by transformer encoding, then various outcomes are classified by the dense layer with GELU and norm. Thus, we obtained the final embedding of text i.e. embedding vector. The BERT utilized the concept of Transformer [11] wherein words' contextual relation in the text is captured using attention mechanism which makes BERT more efficient in feature representation.

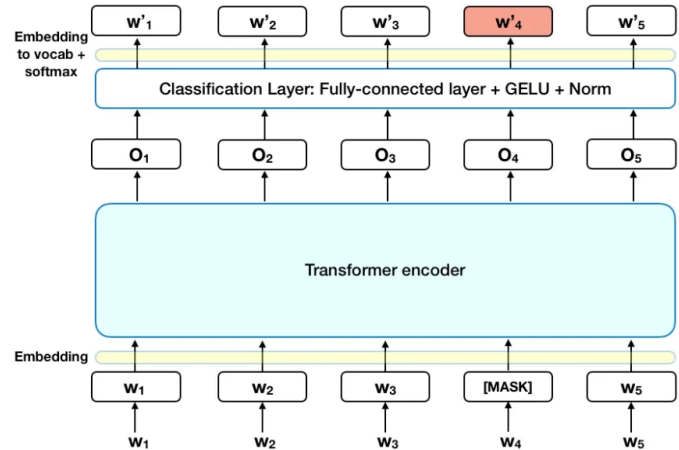

**Figure 3:** BERT Architecture

### 3.5 Hybrid Models

Various techniques exist for constructing hybrid models for SA. In this work, we examined the amalgamation of multiple effective strategies. Proposed methodology commences with the utilization of either GloVe or a pretrained BERT to generate feature vector. Subsequently, we experiment with different arrangements of CNN [46] and LSTM [26] models in subsequent stages followed by the application of Grid Search (GS) for hyperparameter tuning. By incorporating both of these variations, we obtain eight hybrid models that we evaluated for sentiment prediction.

(1) BERT → CNN → Bi-LSTM → Grid Search
(2) BERT → Bi-LSTM → CNN → Grid Search
(3) BERT → CNN → Grid Search
(4) BERT → Bi-LSTM → Grid Search
(5) GloVe → CNN → BiLSTM → Grid Search
(6) GloVe → Bi-LSTM → CNN → Grid Search
(7) GloVe → CNN → Grid Search
(8) GloVe → Bi-LSTM → Grid Search

In this study, we employed a pretrained BERT model with *Layer* = 2 hidden layers, hidden size of *Hidden* = 128, and *Attention* = 2, attention heads. Aforementioned parameter adjustments tune the BERT model to extract features, responsible for producing input data to hybrid models. Data (input) is processed through BERT for creating feature vectors, serving as inputs to the subsequent first four hybrid models designed for classification. Word vectorization in the last four hybrid models is accomplished using GloVe, a technique designed to condense a wealth of textual semantic and grammar information into vectors while effectively decreasing the dimensionality of the vector space. This process enhances the representation of words in a more compact form.

In next step, we combines the CNN and Bi-LSTM models for feature learning based on local spatial features and long-term dependent features for SA, thus, incorporating advantages of these model for efficient SA on the data.

### 3.5.1 CNN

A CNN is a kind of feedforward neural network made up of several layers that analyse and transfer data without cycles in a single direction, from input to output. Its DNN architecture usually starts with layers for convolution and pooling, which change inputs before sending them to a fully linked classification layer. A solo convolutional (1D CNN) is employed in this work. When text processing is utilized for tasks related to NLP - sentiment analysis or text classification, every word in a phrase is frequently represented as a high-dimensional vector. The full sentence is represented by concatenating these word vectors. Zero padding is provided to sentences that have fewer words than the designated length in order to guarantee uniformity in input size. The sentence representation matrix is then subjected to convolutional operations using filters. Each filter, denoted by $W$, has a shape of $h \times k$, where $h$ is the window size (the number of words the filter covers), and $k$ is the dimensionality of the word vectors. The filter is convolved over consecutive windows of words in the sentence representation matrix to produce new feature maps.

For each window of words, represented by $X_{i:i+j}$, a feature vector $C_i$ is generated by Eq. (3).

$$C_i = f(W \times X_{i:i+h-1} + b) \tag{3}$$

Here, $W$ is the filter matrix, $X_{i:i+h-1}$ represents the word's window, $b$ is the bias, and $f$ is a nonlinear activation function like sigmoid or hyperbolic tangent. During training, both the bias term and the filter matrix are learned. The CNN's architecture is presented in Fig. 4.

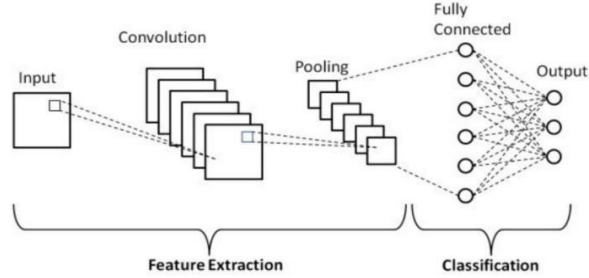

**Figure 4:** Convolutional Neural Network [16].

### 3.5.2 LSTM

Further, LSTM is utilized, as a temporal information extractor which is effective for processing sequential text data, making it suitable for tasks involving forecasting milestones with significant time intervals and time series data that exhibit delays. In LSTM cell, computations of input gate, output gate, and forget gate at time $t$ are given by Eq. (4) to Eq. (9).

$$i_t = \sigma(W^i x_t + U^i h_{t-1}) \tag{4}$$
$$f_t = \sigma(W^f x_t + U^f h_{t-1}) \tag{5}$$
$$o_t = \sigma(W^o x_t + U^o h_{t-1}) \tag{6}$$
$$c'_t = \tanh(W^c x_t + U^c h_{t-1}) \tag{7}$$
$$c_t = i_t \times c'_t + f_t \times c'_{t-1} \tag{8}$$
$$h_t = o_t \times \tanh(c_t) \tag{9}$$

The concatenation of the current input $x_t$ and the prior hidden state $h_{t-1}$ is represented by $[h_{t-1}, x_t]$ in each formula. The weight matrices $W_i$, $W_o$ and $W_f$, along with the bias terms $b_i$, $b_o$ and $b_f$, are parameters learned during training to control the behaviour of the gates in the LSTM network.

However, LSTM has a limitation in that it only considers the text's information in a forward direction, meaning it takes into account the past and present context but not the future. To overcome this limitation, we utilized Bi-LSTM network in proposed model. The Bi-LSTM simultaneously analyzes the textual details in both forward and backward directions. Thus, model incorporates more comprehensive understanding of the context within the text, resulting in improved predictive performance. The adoption of BiLSTM over LSTM, enables the utilization of bidirectional text context information, leading to enhanced prediction performance, especially for tasks involving long intervals and time series data with delays.

Together with input and output blocks and memory cell, Bi-LSTM block also has 3 gates: forget gate, input gate, and output gates. RNNs are good at handling temporal signals, but CNNs excel at handling spatially connected data. Multilayer CNN can adequately collect and learn local information, while Bi-LSTM can retain both forward and backward sequence information. Thus, the combination utilises both the temporal and spatial features to their fullest.

In addition to using CNN and Bi-LSTM in tandem, we also utilized them separately in the sentiment analysis models (1), (2), (5) and (6) with BERT or GloVe as the word embedding to test the performance of individual neural network model. The Bi-LSTM architecture is shown in Fig. 5.

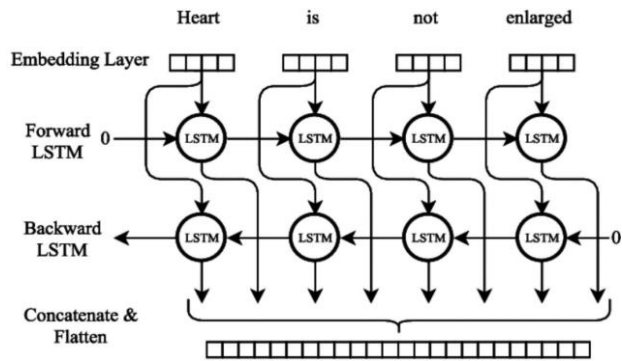

**Figure 5:** Bi-LSTM [17].

### 3.6 Proposed Models

We presented eight hybrid DL models in this part that differ in how GloVe or BERT are utilized in the embedding layer and apply CNN and Bi-LSTM, hybrid models are further discussed in this section.

### 3.6.1 Scenario Combination (1) and (2)

In the first (1) and second (2) model, we leverage BERT as an embedding technique, utilizing its powerful contextual embedding to represent words. In first model, these BERT-generated word vectors are then fed into a CNN, this allow us to capture intricate patterns in the text data using both BERT's contextual understanding and CNN's ability to detect local features. Subsequently, we introduce Bi-LSTM layer to leverage the power of sequential modelling. This sequential aspect further refines the understanding of the text data, capturing dependencies over longer spans. Later, dense layer is applied for classification of sentiment.

In the second model, we introduce a change in the hybrid deep learning model's composition. Here, we apply Bi-LSTM as the initial layer, followed by CNN. This deliberate alteration in the order of layers reflects our exploration of diverse model architectures and their unique capabilities. By employing Bi-LSTM first, we place a stronger emphasis on sequential context before passing the representations through the CNN, thereby seeking to uncover any potential advantages of this rearrangement. This multifaceted approach enriches the understanding of how different deep learning components can be harnessed effectively to solve intricate language-related challenges. We utilized a single layer of CNN, Bi-LSTM, and dense architectures. To improve the overall accuracy of proposed model and cut down on computation time, we used the ReLU activation function in CNN and Bi-LSTM layers and sigmoid activation function in the dense layer.

### 3.6.2 Scenario combination (3) and (4)

In third (3) model, BERT is employed as an embedding technique, and the word vectors are used as input to the CNN, exclusively. In fourth (4) model, we solely apply Bi-LSTM after BERT. Unlike first and second hybrid model where CNN and Bi-LSTM are applied in combination, we applied CNN and Bi-LSTM, individually. Following this, we utilize a grid search algorithm. Combination (3) involved a single layer of CNN and Dense, while in the fourth combination (4), we used a sole layer of Bi-LSTM and Dense. ReLU activation is employed in the CNN and Bi-LSTM layers, further, sigmoid activation enforced in dense layer.

### 3.6.3 Scenario combination (5), (6), (7) and (8)

In fifth (5) model, GloVe-based variant of the first (1) model is utilized, in fifth model, we retain the architecture of the first model while introducing a change in the word embedding technique. Instead of using BERT, we employ GloVe for word embedding. This modification allows us to explore how the performance of the model is affected when relying on pre-trained word vectors from GloVe, which are trained on enormous corpus of textual data. This model serves as a benchmark for comparing the effectiveness of GloVe against BERT in the same architecture.

In sixth (6) model, GloVe-based variant of the second (2) model is used, similar to fifth model, sixth model retains the architecture as in second model but swaps BERT with GloVe. By making this change, we aim to assess how GloVe embedding perform within the context of hybrid model that combines CNN and Bi-LSTM. We are better able to comprehend the relative benefits and drawbacks of each embedding technique in this particular architectures after comparison.

Further, in seventh (7) model, GloVe-based variant of the third (3) model is applied, thus, we maintain the architecture of the third model but replace BERT with GloVe for word embedding. This adjustment enables us to evaluate the impact of using GloVe embedding exclusively in conjunction with a CNN. We'll examine how well GloVe captures contextual information compared to BERT when used in isolation with this architecture. Similarly, in eighth (8) model, GloVe-based variant of the fourth (4) model is utilized, and model retains the structure of the fourth model, with the key distinction being the use of GloVe instead of BERT for word embedding. We may investigate the advantages of integrating GloVe with the Bi-LSTM layer over the contextual embedding of the model with this selection. By comparing this variant to the fourth model, we can assess whether GloVe's strengths align better with sequential modelling in the form of Bi-LSTM.

These eight scenarios enable us to systematically investigate the performance differences between GloVe and BERT as word embedding techniques within various model architectures. This analysis provides valuable insights into the suitability of each embedding method for specific NLP tasks and model configurations. The process of methodology for SA is highlighted in Figure 6. (**Note:** After every layer of CNN, max pooling layer and dropout layer is applied, and after every layer of Bi-LSTM, dropout layer is applied to yield better results.)

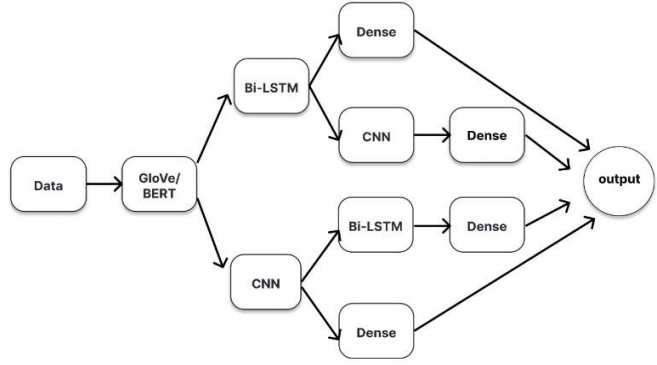

**Figure 6:** Methodology for sentiment analysis.

## 4. Results and Discussion

This section contains the results of experiments we conducted to evaluate the performance of proposed hybrid models. The hyperparameters values are identified after extensive simulation and employing Grid Search algorithm for all the proposed algorithms. The hyperparameters are presented in Table 4 to Table 6. Accuracy and Loss function have been utilised to assess the models' performance throughout every trial. We have used Categorical Cross-Entropy for classification.

Formula of Accuracy:

$$Accuracy = \frac{Total\ Number\ of\ Predictions}{Number\ of\ Correct\ Predictions} \times 100\% \tag{10}$$

Formula of Loss function:

$$Loss = -\frac{1}{N}\sum_{i=1}^{N}\sum_{j=1}^{C} y_{ij} \log(p_{ij}) \tag{11}$$

Where,
$N$ : No. of samples
$C$ : No. of classes
$y_{ij}$ : Ground truth label for the $i^{th}$ sample and the $j^{th}$ class
$p_{ij}$ : Predicted probability that the $i^{th}$ sample belongs to the $j^{th}$ class

**Table 4:** Grid Search Parameters for scenario 1, 2, 5, 6.

| Parameter Name | Parameter Values |
|---|---|
| Filter 1 (1st Layer) | 128, 256, and 512 |
| Filter 2 (2nd Layer) | 64, 128, 256, and 512 |
| batch size | 128, 256, and 512 |

**Table 5:** Grid Search Parameters for scenario 3, 4, 7, 8.

| Parameter Name | Parameter Values |
|---|---|
| Filter 1 (1st Layer) | 64, 128, 256, and 512 |
| batch size | 128, 256, and 512 |

**Table 6:** Hyperparameters values.

| Hyperparameters | Values |
|---|---|
| Optimizer | Adam |
| Filter Size | 10 |
| Echoes | 4 |
| Learning rate | 0.001 |
| Dropout | 0.5 |

Before conducting the experiments, we carefully configured various parameters, hardware components, and essential library facilities. Experiments are conducted on Online Kaggle platform with GPU P100, 16GB and Keras [38] and TensorFlow [37] libraries. Throughout the experiments, we consistently set the parameters, including 'echoes' to 4 and 'k-fold' to 10.

The K-fold cross-validation is commonly utilized method in ML to assess model performance, typically using values of $k$ equal to 3, 5, or 10. Among these, $k = 10$ is the most prevalent choice, especially when dealing with large datasets. Consequently, we divided dataset into three parts, where 2 parts of training and one part of testing, the cross validation to 3, while applying Grid Search. The choice of $k = 10$ ensures that each training and test sample is sufficiently representative of the dataset. But, this method ensures that the $k$ test sets in every iteration are same size, and that all $k$ models in cross-validation are trained on datasets of same size. We have applied padding to be same in every layer to ensure the uniformity. The performance evaluation is methodically presented in Table 7 for Model (1), Model (2), Model (5) and Model (6), and Table 8 shows the performance of Models (3), (4), (7), and (8). Further, Tables 9 shows the performance of eight proposed hybrid models with higher accuracy. Table 10 presents the comparative analysis on the Twitter COVID-19 dataset, we observe the highest performance achieved by the proposed models in comparison with the existing models. Table 11 highlights the evaluation on different datasets for the sentiment analysis, and proposed models outperforms the state-of-the-arts, highlighting the significance of the best hyperparameter configurations and providing information on relative advantages and disadvantages of each strategy in the context of the research. The results presented in Table 7 to Table 11 offer a comprehensive overview of the models' effectiveness, facilitating a comparative assessment of their performance based on diverse hyperparameter settings.

**Table 7:** Accuracy and loss of model (1), (2), (5), (6).

| Parameters | | | Model 1 | | Model 2 | | Model 5 | | Model 6 | |
|---|---|---|---|---|---|---|---|---|---|---|
| | | | BERT->CNN->Bi-LSTM | | BERT->Bi-LSTM->CNN | | GloVe->CNN->Bi-LSTM | | GloVe->BiLSTM>CNN | |
| Batch Size | Filter 1 | Filter 2 | Accuracy | Loss | Accuracy | Loss | Accuracy | Loss | Accuracy | Loss |
| 128 | 128 | 64 | 0.934076 | 0.00378 | 0.933184 | 0.002763 | 0.976823 | 0.00059 | 0.977925 | 0.00247 |
| 128 | 128 | 128 | 0.93813 | 0.003419 | 0.932654 | 0.004937 | 0.976949 | 0.000786 | 0.980498 | 0.00129 |
| 128 | 128 | 256 | 0.942193 | 0.003418 | 0.94179 | 0.00204 | 0.97577 | 0.001534 | 0.978908 | 0.00106 |
| 128 | 128 | 512 | 0.94501 | 0.003951 | 0.937098 | 0.005205 | 0.976147 | 0.001019 | 0.979584 | 0.000962 |
| 128 | 256 | 64 | 0.931492 | 0.003044 | 0.931771 | 0.008591 | 0.976649 | 0.00041 | 0.981202 | 0.000633 |
| 128 | 256 | 128 | 0.939599 | 0.002515 | 0.945614 | 0.003787 | 0.975247 | 0.001954 | **0.981795** | **0.001298** |
| 128 | 256 | 256 | 0.934532 | 0.003662 | 0.94924 | 0.001791 | 0.974752 | 0.003079 | 0.981118 | 0.000677 |
| 128 | 256 | 512 | 0.938632 | 0.005002 | 0.951266 | 0.003792 | **0.976977** | **0.000316** | 0.980526 | 0.001171 |
| 128 | 512 | 64 | 0.90151 | 0.021745 | 0.925505 | 0.007105 | 0.975038 | 0.000773 | 0.978873 | 0.002086 |
| 128 | 512 | 128 | 0.929177 | 0.00148 | 0.939236 | 0.003382 | 0.975903 | 0.002123 | 0.978873 | 0.000571 |
| 128 | 512 | 256 | 0.928136 | 0.006688 | 0.945632 | 0.001531 | 0.975756 | 0.000784 | 0.981488 | 0.001247 |
| 128 | 512 | 512 | 0.930395 | 0.005469 | 0.949472 | 0.00235 | 0.975589 | 0.001294 | 0.980993 | 0.000936 |
| 256 | 128 | 64 | 0.947882 | 0.005317 | 0.945335 | 0.004075 | 0.97538 | 0.000123 | 0.977716 | 0.000677 |
| 256 | 128 | 128 | 0.95716 | 0.001801 | 0.956147 | 0.003622 | 0.975003 | 0.000444 | 0.978266 | 0.000089 |
| 256 | 128 | 256 | 0.952912 | 0.005951 | 0.962218 | 0.003533 | 0.975401 | 0.001411 | 0.978057 | 0.000402 |
| 256 | 128 | 512 | 0.961595 | 0.001267 | 0.961995 | 0.002012 | 0.974989 | 0.000355 | 0.976893 | 0.001624 |
| 256 | 256 | 64 | 0.948942 | 0.005652 | 0.94963 | 0.003818 | 0.975721 | 0.000401 | 0.979863 | 0.000231 |
| 256 | 256 | 128 | 0.958992 | 0.001965 | 0.954743 | 0.001935 | 0.975136 | 0.000278 | 0.979828 | 0.000924 |
| 256 | 256 | 256 | 0.946571 | 0.004854 | 0.95651 | 0.004155 | 0.976021 | 0.000245 | 0.97934 | 0.000823 |
| 256 | 256 | 512 | 0.951815 | 0.003714 | 0.953311 | 0.00443 | 0.975561 | 0.000547 | 0.978448 | 0.001126 |
| 256 | 512 | 64 | 0.949063 | 0.005708 | 0.945698 | 0.003816 | 0.973971 | 0.001923 | 0.981237 | 0.000546 |
| 256 | 512 | 128 | 0.957067 | 0.002001 | 0.952224 | 0.006234 | 0.975617 | 0.000444 | 0.978769 | 0.001125 |
| 256 | 512 | 256 | 0.945456 | 0.003564 | 0.959364 | 0.000991 | 0.974697 | 0.000849 | 0.977674 | 0.004351 |
| 256 | 512 | 512 | 0.950904 | 0.002972 | 0.961456 | 0.0018 | 0.975617 | 0.000513 | 0.979103 | 0.001982 |
| 512 | 128 | 64 | 0.963808 | 0.01235 | 0.961604 | 0.011265 | 0.972765 | 0.001209 | 0.974606 | 0.000418 |
| 512 | 128 | 128 | 0.981992 | 0.004156 | 0.977827 | 0.001793 | 0.974111 | 0.00039 | 0.975324 | 0.001004 |
| 512 | 128 | 256 | 0.962664 | 0.013824 | 0.962943 | 0.012309 | 0.972549 | 0.000863 | 0.974515 | 0.000397 |
| 512 | 128 | 512 | **0.983768** | **0.003837** | 0.979835 | 0.003457 | 0.972207 | 0.002182 | 0.973937 | 0.001327 |
| 512 | 256 | 64 | 0.962655 | 0.01148 | 0.957021 | 0.011854 | 0.974125 | 0.000633 | 0.97773 | 0.000365 |
| 512 | 256 | 128 | 0.98123 | 0.00333 | 0.97528 | 0.004053 | 0.973441 | 0.000849 | 0.976014 | 0.000291 |
| 512 | 256 | 256 | 0.960442 | 0.010569 | 0.963352 | 0.014417 | 0.97425 | 0.00073 | 0.976481 | 0.001169 |
| 512 | 256 | 512 | 0.954093 | 0.018411 | **0.982996** | **0.005322** | 0.973818 | 0.0002 | 0.976098 | 0.000923 |
| 512 | 512 | 64 | 0.958834 | 0.010677 | 0.956026 | 0.01312 | 0.972396 | 0.000835 | 0.97697 | 0.000611 |
| 512 | 512 | 128 | 0.973876 | 0.003973 | 0.977753 | 0.006071 | 0.972298 | 0.001048 | 0.977242 | 0.000772 |
| 512 | 512 | 256 | 0.95773 | 0.000365 | 0.958127 | 0.009837 | 0.973985 | 0.000419 | 0.978001 | 0.001319 |
| 512 | 512 | 512 | 0.976014 | 0.000291 | 0.971403 | 0.00289 | 0.973365 | 0.000084 | 0.978301 | 0.000286 |

**Table 8:** Accuracy loss of model (3), (4), (7), (8).

| Parameters | | Model 3 | | Model 4 | | Model 7 | | Model 8 | |
|---|---|---|---|---|---|---|---|---|---|
| | | BERT->CNN | | BERT->Bi-LSTM | | GloVe->CNN | | GloVe->Bi-LSTM | |
| Batch Size | Filter 1 | Accuracy | Loss | Accuracy | Loss | Accuracy | Loss | Accuracy | Loss |
| 128 | 64 | 0.953144 | 0.009493 | 0.957895 | 0.008902 | **0.974522** | **0.001109** | 0.979529 | 0.000975 |
| 128 | 128 | 0.953851 | 0.006149 | 0.972426 | 0.003309 | 0.972702 | 0.001761 | 0.980044 | 0.001946 |
| 128 | 256 | 0.961809 | 0.006125 | 0.944954 | 0.003137 | 0.973713 | 0.00079 | **0.980839** | **0.000857** |
| 128 | 512 | 0.966848 | 0.004344 | 0.951025 | 0.004196 | 0.973177 | 0.000247 | 0.980602 | 0.002234 |
| 256 | 64 | 0.954551 | 0.005902 | 0.957895 | 0.008902 | 0.97448 | 0.000291 | 0.977032 | 0.00232 |
| 256 | 128 | 0.978765 | 0.003709 | 0.972426 | 0.003309 | 0.973755 | 0.001369 | 0.979222 | 0.000346 |
| 256 | 256 | 0.9575 | 0.009818 | 0.955589 | 0.009818 | 0.973964 | 0.000351 | 0.979466 | 0.000899 |
| 256 | 512 | 0.977898 | 0.001924 | 0.973077 | 0.001424 | 0.973044 | 0.001002 | 0.979229 | 0.000112 |
| 512 | 64 | 0.962869 | 0.011888 | 0.96272 | 0.013428 | 0.973051 | 0.00056 | 0.975254 | 0.000621 |
| 512 | 128 | 0.986008 | 0.002788 | **0.985274** | **0.004067** | 0.973466 | 0.000304 | 0.975966 | 0.000995 |
| 512 | 256 | 0.965862 | 0.013076 | 0.963594 | 0.011927 | 0.973553 | 0.000552 | 0.976391 | 0.000809 |

|  |  | 512 | 512 | **0.988658** | **0.002802** | 0.984874 | 0.00409 | 0.973274 | 0.000129 | 0.97849 | 0.00047 |

**Table 9:** Accuracy obtained from proposed models on Twitter COVID-19 dataset.

| Model No. | Hybrid model | Accuracy (%) |
|---|---|---|
| Model 1 | BERT + CNN + Bi-LSTM + GS | 0.983768 |
| Model 2 | BERT + Bi-LSTM + CNN + GS | 0.982996 |
| Model 3 | BERT + CNN + GS | **0.988658** |
| Model 4 | BERT + Bi-LSTM + GS | 0.985274 |
| Model 5 | GloVe + CNN + Bi-LSTM + GS | 0.976977 |
| Model 6 | Glove + Bi-LSTM + CNN + GS | 0.981795 |
| Model 7 | Glove + CNN + GS | 0.974522 |
| Model 8 | Glove + Bi-LSTM + GS | 0.980839 |

**Table 10:** Comparison of proposed models with state-of-the-arts on Twitter COVID-19 dataset.

| Ref. | Model | Accuracy (%) |
|---|---|---|
| Proposed Model | BERT + CNN + GS | 98.86 |
| Xiaoyan et al. [9] | GloVe + CNN + BiLSTM | 95.65 |
| Ainapure et al. [17] | Bi-LSTM | 94.48 |
| Qorib et al. [5] | TextBlog + TF-IDF + LinearSVC | 96.75 |
| Topbaş et al. [18] | BERT | 83.14 |
| Topbaş et al. [18] | RNN | 86.40 |
| Chakraborty et al. [15] | LSTM | 84.46 |
| Chakraborty et al. [47] | CNN + LSTM | 86.58 |
|  | CNN + Bi-LSTM | 87.22 |

Table 10 shows the highest accuracy of 98.86% for proposed BERT + CNN + GS. The obtained result is 2.11% higher in terms of accuracy in comparison with model composition presented by Qorib et al. [5]. In [5], sentiment classification is accomplished by using TextBlog, TF-IDF and different ML algorithms, and TextBlog + TF-IDF + LinearSVC offers good results (i.e. 96.75%). So, the increase in accuracy is reported for BERT and CNN along with GS. This is due to changing the word-embedding and adopting CNN, enhanced the performance of model. In addition, batch size (512) and filter size (512) analysis shows the better hyperparameter tuning for enhanced accuracy. Moreover, we have addressed the future challenges listed by Xiaoyan et al. [9] by developing hybrid model using BERT, CNN and Bi-LSTM wherein highest accuracy is achieved in comparison with the latest work on sentiment prediction. In Xiaoyan et al. [9], GloVe + CNN + Bi-LSTM is experimented on different length tweets and achieved accuracy of 95.65%, but with same combination and employing Grid Search algorithm enhances the accuracy by 2.04% by the proposed model. Further, improved accuracy of 3.21% is obtained for BERT + CNN + GS, herein, change in embedding and Grid Search archives the increased accuracy.

In Topbaş et al. [18], single model of Bi-LSTM or GRU is adopted for tweets analysis and obtained the accuracy of 94.48% which is 4.05% and 4.38% lower accuracy than BERT + Bi-LSTM + GS and BERT + CNN + GS models, respectively. This is due to the impact of word-embedding and GS with DL model.

Further, an improved accuracy of 12.48% and 15.72% is identified by comparing [19] wherein direct utilization of RNN and BERT is investigated. So, we conclude that the use of hybrid model with varied word-embedding offers improved accuracy for sentiment classification on tweets. Lastly, authors in [15], achieved the accuracy of 84.46% by employing BoW + LSTM which is 14.40% lower in comparison with BERT + CNN + GS, this is due to better feature representation and suitable hyperparameter tuning.

Table 11 shows that proposed model outperforms state-of-the-arts on various datasets, this facts can be utilized for the experimentation of same models on different datasets to achieve the model generalization. In future, this will be explore on different datasets to observe the similar results.

**Table 11:** Comparison of proposed model with different models on different dataset.

| Ref. | Model | Dataset | Accuracy (%) |
|---|---|---|---|
| Proposed model | BERT + CNN + GS | Twitter Covid-19 | 98.86 |
| Qorib et al. [5] | TextBlog + TF-IDF + LinearSVC | Twitter Covid-19 | 96.75 |
| Xiaoyan et al. [9] | GloVe + CNN + BiLSTM | Twitter Covid-19 | 95.65 |
| Priyadarshini et al. [10] | LSTM + CNN + GS | Amazon Reviews | 96.40 |
| Yang et al. [12] | XLNET | IMDB | 96.21 |
| Dang et al. [14] | BERT + CNN + LSTM | IMDB | 93.40 |
| Alaparthi ans Mishra [36] | BERT | IMDB | 92.31 |

The summarization of key findings from the experiments are as follows:

1. GloVe *vs* BERT embedding: We observed that hybrid models utilizing BERT as the embedding technique achieved higher accuracy in sentiment analysis compared to GloVe. Additionally, it is worth noting that the computation time required for GloVe models is significantly shorter than the BERT models.
2. Leveraging CNN and Bi-LSTM: Combination of CNN and Bi-LSTM as hybrid models proved beneficial. CNN excelled in extracting essential local characteristics from the data, while Bi-LSTM effectively stored both past and future information at the state nodes (i.e. cell state). This combination allowed us to leverage the strengths of both architectures. But, while utilizing BERT models, single model utilization giving better results.
3. Grid Search for Hyperparameter Tuning: The use of Grid Search in classification method yields improved results. It provided the detailed understanding of how different hyperparameters influenced the performance of hybrid models. This approach allowed us to fine-tune the models for efficient sentiment prediction.

Thus, experiments showcased the benefits of hybrid DL models, highlighted effectiveness of BERT embedding over GloVe, and emphasized the synergy between CNN and Bi-LSTM in sentiment analysis. Furthermore, using Grid Search to adjust hyperparameters improved comprehension of the hybrid model's functionality. These findings contribute to the broader understanding of sentiment analysis on social network data.

Furthermore, due to insufficient computational resources, we were unable to integrate the BERT LARGE model into proposed methodology. We believe, the utilizing of BERT LARGE into the proposed model could potentially improve the results.

## 5. Conclusions

Due to COVID-19 vaccination drives, peoples were reluctant for vaccination due to various myths and health hazards wherein acceptance was reported by some group of peoples. The researchers investigating COVID-19 combat after vaccination, and the varied opinions are reported on social media. Thus, individual sentiment on COVID-19 tweets attracted the attention of different agencies and researchers. The tweets analysis from Twitter COVID-19 dataset shows the inclination towards COVID-19 vaccine, which shows the encouraging attitude of society for vaccination drives. Further, we observe the increase in positive opinion and reduction in negative or neutral opinions. Even, individuals coming forward for booster dosage of COVID-19 vaccine.

To analyse the sentiments, we propose hybrid DL models on Twitter COVID-19 dataset. These models combines CNN, Bi-LSTM models, and utilizing two distinct word embedding techniques: GloVe and BERT. We conducted studies using Twitter datasets that included COVID-19-related messages. The primary objective is to assess the performance and adaptability of hybrid models and model's ability to analyze sentiment across a diverse range of tweets. Additionally, we looked into how different feature extraction methods and DL models affected the accuracy of sentiment polarity. The outcomes show that models using GloVe are consistently outperformed by CNN and Bi-LSTM with BERT. The highest classification accuracy of 98.86% is achieved for BERT + CNN + GS in comparison with existing works. Further, this model is less complex compared to Glove + BiLSTM + CNN + GS wherein 98.17% accuracy is reported which is 0.69% lesser to the best performing model. Thus, we suggest the use of BERT + CNN + GS for twitter sentiment analysis with optimal hyperparameters.

**Declarations**

**Conflict of interest**

There is no conflict of interest.

**Funding**

No funding used for this work.

**Data availability**

The dataset is publically available, which is mentioned in this paper.